\documentclass[a4paper, 10pt, conference]{ieeeconf}      
\usepackage{FG2019}

\FGfinalcopy 

\IEEEoverridecommandlockouts                              
\usepackage{graphics} 
\usepackage{epsfig} 
\usepackage{mathptmx} 
\usepackage{times} 
\usepackage{amsmath} 
\usepackage{amssymb}  
\usepackage{subfigure}
\usepackage{graphicx}
\usepackage{adjustbox}
\usepackage{hhline}
\usepackage{tabularx}
\usepackage{verbatim}
\usepackage{amsmath}

\usepackage{pgffor}
\usepackage{caption}

\def\FGPaperID{198} 

\title{\LARGE \bf
A Multi-Task Learning \& Generation Framework: Valence-Arousal, Action Units \&  Primary Expressions
}

\begin{document}

\ifFGfinal

\author{\parbox{16cm}{\centering
    {\large Dimitrios Kollias$^1$ and Stefanos Zafeiriou$^{1,2}$}\\ 
    {\normalsize
    $^1$  Department of Computing, Imperial College London, UK \\
    $^2$ Centre for Machine Vision and Signal Analysis, University of Oulu, Finland}
    }
}
\thispagestyle{empty}
\pagestyle{empty}
\fi
\maketitle

\begin{abstract} 
Over the past few years many research efforts have been devoted to the field of affect analysis. Various approaches have been proposed for: i) discrete emotion recognition in terms of the primary facial expressions; ii) emotion analysis in terms of facial Action Units (AUs), assuming a fixed expression intensity; iii) dimensional emotion analysis, in terms of valence and arousal (VA). These approaches can only be effective, if they are developed using large, appropriately annotated databases, showing behaviors of people in-the-wild, i.e., in uncontrolled environments. Aff-Wild  has been the first, large-scale, in-the-wild database (including around 1,200,000 frames of 300 videos), annotated in terms of VA. In the vast majority of existing emotion databases, their annotation is limited to either primary expressions, or valence-arousal, or action units. In this paper, we first annotate a part (around $234,000$ frames) of the Aff-Wild database in terms of $8$ AUs and another part (around $288,000$ frames) in terms of the $7$ basic emotion categories, so that parts of this database are annotated in terms of VA, as well as AUs, or primary expressions. Then, we set up and tackle multi-task learning for emotion recognition, as well as for facial image generation. Multi-task learning is performed using: i) a deep neural network with shared hidden
layers, which learns emotional attributes by exploiting their inter-dependencies; ii) a discriminator of a generative adversarial network (GAN). On the other hand, image generation is implemented through the generator of the GAN. For these two tasks, we carefully design loss functions that fit the examined set-up. Experiments are presented which illustrate the good performance of the proposed approach when applied to the new  annotated parts of the Aff-Wild database.

\end{abstract}

\section{INTRODUCTION}

\begin{figure}[h]
\centering
\adjincludegraphics[height=4.5cm,width=8.5cm]{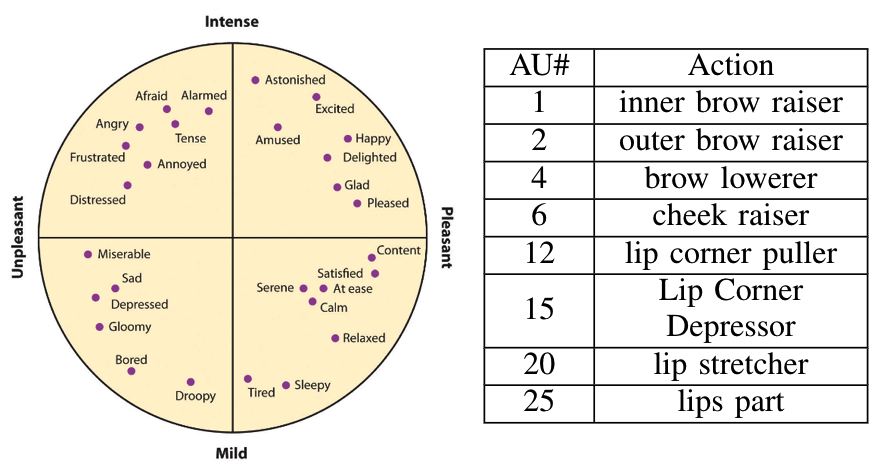}
\caption{The 2D Emotion Wheel (left hand side); the most common AUs with their definitions (right hand side) }
\label{2d-wheel-au}
\vskip -0.5cm
\end{figure}

Representing human emotions has been a basic topic of research in psychology. The most frequently used emotion representation 
is the categorical one, including the seven basic categories, i.e., Anger, Disgust, Fear, Happiness, Sadness, Surprise and Neutral \cite{dalgleish2000handbook}\cite{cowie2003describing}. Discrete emotion representation can also be described in terms of the 
Facial Action Coding System (FACS) model, in which all possible facial actions are described in terms of Action
Units (AUs) \cite{ekman1977facial}. Finally, the dimensional model of affect \cite{whissel1989dictionary}\cite{russell1978evidence} has been proposed as a means to distinguish between subtly different displays of affect and encode small changes in the intensity of each emotion on a continuous scale. The 2-D Valence and Arousal Space (VA-Space) is the most usual dimensional emotion representation. Figure \ref{2d-wheel-au} shows: i) on the left hand side, the 2-D Emotion Wheel \cite{plutchik1980emotion}, with valence ranging from very positive to very negative and arousal from very active to very passive; ii) on the right hand side, some of the most common AUs along with definitions of the represented actions. 

Automatic understanding of human affect using visual signals is a problem that has attracted significant interest over the past 20 years, in many application areas, such as medicine \cite{tagaris1,tagaris2}, health \cite{kollias13} and entertainment.
Current research in automatic analysis of facial affect aims at developing systems, such as robots and virtual humans, that will interact with humans in a naturalistic way under real-world settings \cite{kollias10,kollias11}. To this end, such systems should automatically sense and interpret facial signals relevant to emotions, appraisals and intentions. 

Basic research in face perception and emotion theory cannot be completed without large annotated databases of
images and video sequences of facial expressions and underlying emotions.
Some datasets that have been developed in the labs and are still used in many recent works include the Cohn-Kanade database \cite{tian2001recognizing}\cite{lucey2010extended}, MMI database  \cite{pantic2005web}\cite{valstar2010induced}, Multi-PIE database \cite{gross2010multi} and BU-3D/BU-4D ones \cite{yin20063d}\cite{yin2008high}. 

Previous studies have reported good results in the automatic analysis of facial expressions and related
emotions \cite{corneanu2016survey}. However, these results were obtained with
analysis of images and videos captured in laboratory environments. That is, even when the expressions were spontaneous,
the filming was done in controlled conditions, with full awareness of the participants.

Hence, efforts have been made in order to collect videos of subjects displaying behaviors in-the-wild. To this end, Aff-Wild was created \cite{kollias1,kollias3}, constituting the first large-scale "in-the-wild" database, with over 60 hours of video data, annotated in terms of valence-arousal dimensions. However, even in this case,  annotation is limited only to a single emotion representation, i.e., the VA one. Generating databases which are annotated in terms of more than a single  emotion representation could assist in developing domain adaptation \cite{kollias6,kollias12} and image generation techniques \cite{kollias8,kollias9}, as well as multi-task learning.
Deep generative models have become widely popular for generative modeling of data. Generative adversarial networks (GANs) \cite{goodfellow2014generative}, in particular, have shown remarkable success in generating very realistic images in several cases \cite{radford2015unsupervised}\cite{berthelot2017began}. A typical GAN consists of the discriminator - which tries to tell apart real from fake examples by minimizing an appropriate loss function - and the generator - which tries to generate samples that maximize that loss \cite{tu2007learning}.

One of the primary motivations for studying deep generative models has been semi-supervised learning.
Indeed, several recent works have shown promising empirical results on semi-supervised learning with GANs. Most state-of-the-art semi-supervised learning methods based on GANs  \cite{salimans2016improved} use the GAN discriminator as a classifier which outputs $k+1$ probabilities ($k$ probabilities for the $k$ real classes and one for the fake class). The generator is mainly used as a source of additional data (fake samples) which the discriminator tries to classify under the $(k+1)$th label.

Multi-task learning (MTL) \cite{caruana1997multitask} is a machine learning paradigm for learning a number of supervised learning tasks simultaneously, exploiting commonalities between them. MTL proved to successfully boost the performance of an individual task with the inclusion of other correlated tasks in the training process \cite{ganin2014unsupervised}\cite{hinton2015distilling}.
MTL was first studied in \cite{caruana1997multitask}, where the authors proposed to jointly learn parallel tasks sharing a common representation, and transferring part of the knowledge learned to solve one task to improve the learning of the other related tasks. 
One of the main difficulties with multitask approaches using different databases is the fact that not all the samples are labeled for all the tasks.

In this work we make the following contributions: \\ 
1) We annotate a part of the Aff-Wild database in terms of eight AUs and another part in terms of the seven basic expressions, exploiting its in-the-wild nature with:  i) great variability of behaviors, ii) wide range of emotions, iii) rapid emotional changes and iv) different head poses, illumination conditions and occlusions. This work is a preliminary of \cite{kollias4,kollias15}.\\
2) By adapting current state-of-the-art GAN architectures to semi-supervised settings and by using the above annotations, we generate: i) realistic and vivid images of the persons appearing in the newly annotated parts of Aff-Wild and ii) new images of either unseen people, or new expressions and features of people already appearing in them.\\
3) By designing and testing new appropriate loss functions for our data, we perform MTL experiments: i) using CNN-RNN networks with shared hidden
layers, that jointly learn emotional attributes by exploiting their inter-dependencies and ii) using the GAN described in (2) above, so that it can generate realistic images, whilst serving as a good classifier and regressor.

\section{The Multi-labelled  Aff-Wild parts}

\subsection{The existing Aff-Wild annotated for  Valence \& Arousal}

The Aff-Wild was the first, large scale, in-the-wild, database, consisting of 298 videos and displaying reactions
of 200 subjects, with a total video duration of more than 30 hours. The total number of frames of this database was 1,180,000. The total number of subjects was 200, with 130 of them being male and 70 of them female.
This database has been annotated by 8 lay experts with regards to valence and arousal (VA). The Aff-Wild database served as benchmark for the Aff-Wild Challenge, organized in conjunction with CVPR 2017. The aim for this database was to collect spontaneous facial behaviors in arbitrary recording conditions. To
this end, the videos were collected by searching the Youtube video sharing web-site, mainly with the "reaction" keyword.

\subsection{The Aff-Wild part annotated for eight Action Units}\label{action_unit_section}

We carefully selected $64$ videos from the Aff-Wild database with a total length of $2$ hours and $10$ mins. All the videos were in MP4 format. The videos showed people being active and doing facial movements, thus leading to AU activation. 

 The selected 64 videos had $234,000$ number of frames. They contained 64  subjects, with 40 of them being male and 24 female. Table \ref{attrs_au} shows the attributes of the annotated, in terms of AUs, part of Aff-Wild.  

\begin{table}[h]
\caption{Attributes of the AU annotated part of Aff-Wild}
\label{attrs_au}
\centering
\begin{tabular}{|c|c|}
\hline
AU \# & Description \\ 
\hhline{|=|=|}
No of frames  & $234,000$ \\
\hline
No of videos & $64$ \\
\hline
No of subjects  & $64$ ($40$ male; $24$ female)  \\
\hline
No of annotators  & $3$ \\
\hline
Length of videos & $3$ secs $-$ $16$ mins $21$ secs  \\
\hline
Video format & MP$4$\\
\hline
Mean Image Resolution & $827 \times 516 $\\
\hline
\end{tabular}
\end{table}

The annotation was performed with respect to AU 1, 2, 4, 6, 12, 15, 20, 25, that are shown in Figure \ref{2d-wheel-au}. Three experts annotated those videos. Table \ref{table:ann_1} shows some images with their corresponding VA and AU annotations.

\begin{table}[h]
\caption{ Images with their corresponding VA and AU annotations}
\label{table:ann_1}
\centering
\scalebox{1}{
\begin{tabular}[scale=1]{|c|c|c|c|c|}
 \hline
 \multicolumn{1}{|c|}{Annotation} & \multicolumn{4}{c|}{Images}  \\

			\hline
			 & a & b & c & d   \\
            \hline
              &
 \includegraphics[width=1.2cm]{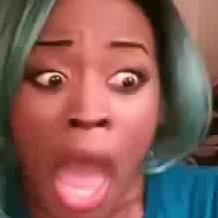} &   \includegraphics[width=1.2cm]{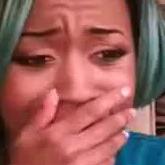} &
  \includegraphics[width=1.2cm]{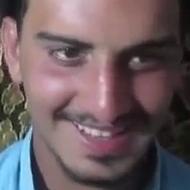} &  \includegraphics[width=1.2cm]{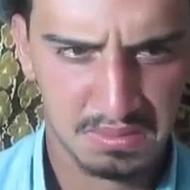}
 \\
  
            \hline
            Valence  & -0.69 & -0.54 & 0.38 & -0.30  \\
            Arousal   & 0.92 & 0.52 & 0.35 & 0.51 \\
            AU 1  &  x & &  &  \\
            AU 2  &  & & & \\
            AU 4  &  & x &  & x \\
            AU 6 &  & x &  & \\
            AU 12    & & & x & \\
            AU 15   & & & & x \\
            AU 20   & & &  & \\
            AU 25   & x & & x & \\
            \hline
		\end{tabular}}
\end{table}

From those 64 videos, the total number of frames that contained at least one of the above AUs was $139,298$. Table \ref{tab:frame_number_action_unit} shows the AU distribution in the annotated Aff-Wild part. 

\begin{table}[h]
    \centering
        \caption{Total number of frames with a specific AU}
    \label{tab:frame_number_action_unit}
\begin{tabular}{|c|c|c|c|}\hline
  Action Unit \# & \begin{tabular}{@{}c@{}}Total Number \\ of Frames \end{tabular} & Action Unit \# & \begin{tabular}{@{}c@{}}Total Number \\ of Frames \end{tabular} \\
  \hline
   AU 1 & 43,948  
   & AU 2  & 25,312 \\
   \hline
   AU 4 & 38,879  
&
   AU 6  & 48,185  \\
   \hline
   AU 12  & 45,291 
&
   AU 15 & 7,023  \\
      \hline
   AU 20 & 9,270  
&   AU 25 & 17,741 \\
   \hline
\end{tabular}
\end{table}

\begin{figure}[h]
\centering
\adjincludegraphics[height=4.4cm,width=6.9cm]{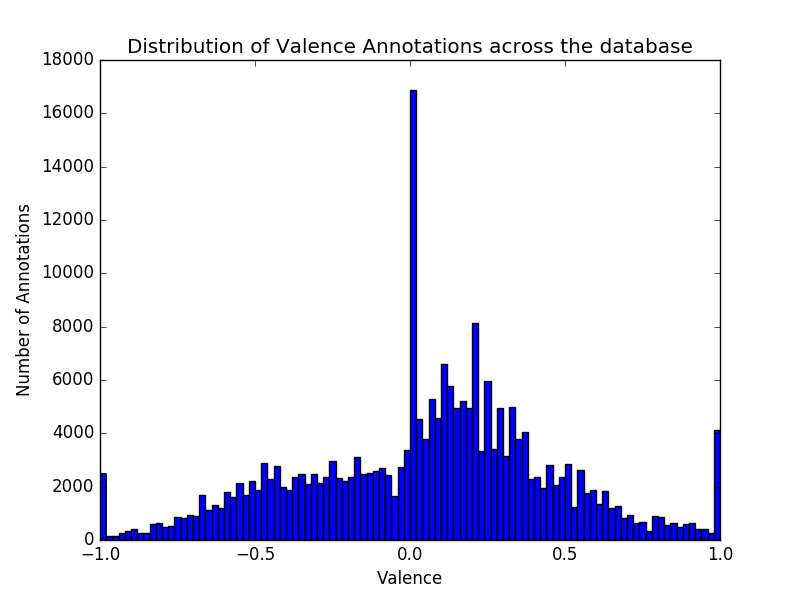}\\
\adjincludegraphics[height=4.4cm,width=6.9cm]{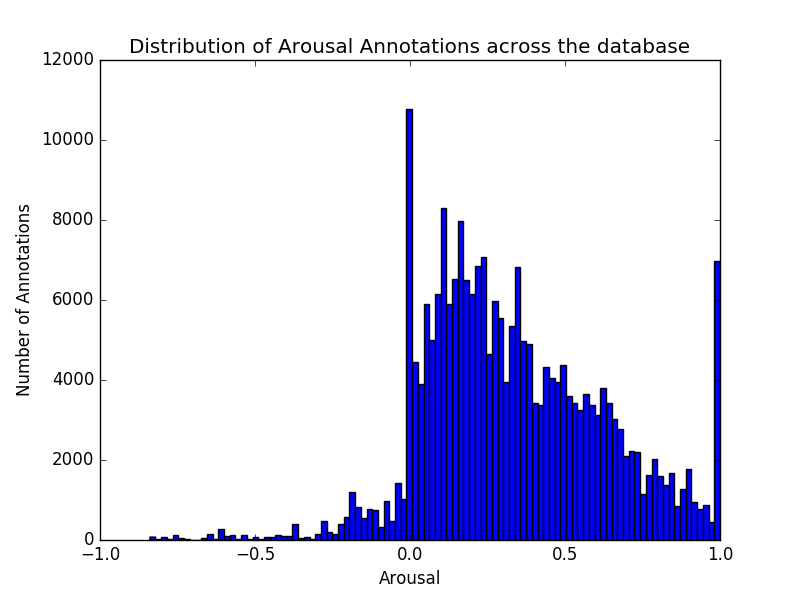}
\caption{Histogram of valence and arousal annotations of the part of Aff-Wild database that was annotated for AU.}
\vskip -0.5cm
\label{hist_va_au}
\end{figure}

Figure \ref{hist_va_au} provides a histogram of the valence and arousal
values in the annotated part of Aff-Wild. 

\subsection{The part of Aff-Wild annotated for the basic expressions}

We carefully selected $55$ videos from the Aff-Wild database with a total length of $2$ hours and $40$ mins. All the videos were in MP4 format.
Aff-Wild contains both subtle and extreme human behaviours in real-world settings. Due to the complexity of such behaviours, the seven basic expressions and underlying emotions are not so commonly displayed in them. 
 

Nevertheless, $55$ videos were found and selected, consisting of $288,000$ number of frames. They contained 56 subjects, with $25$ of them being male and $31$ female. Table \ref{attrs2} shows the attributes of the annotated part of Aff-Wild.

\begin{table}[h]
\caption{Attributes of the annotated part of Aff-Wild for the basic expressions}
\label{attrs2}
\centering
\begin{tabular}{|c|c|}
\hline
Basic Expressions \# & Description \\ 
\hhline{|=|=|}
No of frames  & $288,000$ \\
\hline
No of videos & $55$ \\
\hline
No of subjects  & $56$ ($25$ male; $31$ female)  \\
\hline
No of annotators  & $3$ \\
\hline
Length of videos & $4$ secs $-$ $26$ mins $22$ secs  \\
\hline
Video format & MP$4$\\
\hline
Mean Image Resolution & $1297 \times 775 $\\
\hline
\end{tabular}
\end{table}

\begin{figure}[h]
\centering
\includegraphics[height=4.2cm,width=6cm]{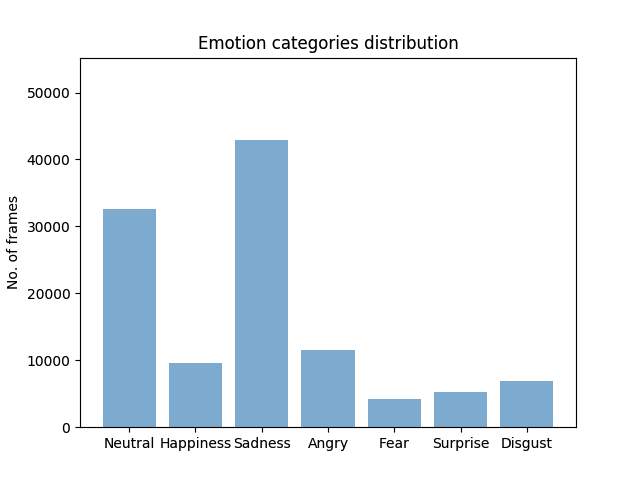}
\caption{Histogram of the seven basic expressions in the annotated part of Aff-Wild}
\label{tab:frame_number_basic_expr}
\vskip -0.2cm
\end{figure}

The same three experts, who annotated a part of the Aff-Wild in terms of AUs in Section \ref{action_unit_section}, annotated the above videos in terms of the seven basic expressions, as well.
From those 55 videos, the total number of frames that contained at least one of the seven basic expressions was $115,640$. Figure \ref{tab:frame_number_basic_expr} shows the histogram of the seven basic expressions in the annotated part.

\begin{figure}[h]
\centering
\adjincludegraphics[height=4.5cm,width=6.5cm]{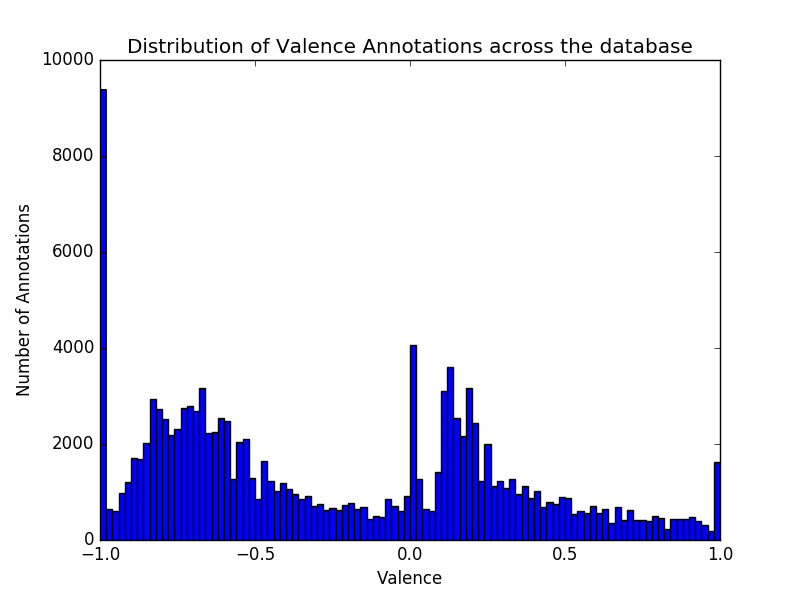}\\
\adjincludegraphics[height=4.5cm,width=6.5cm]{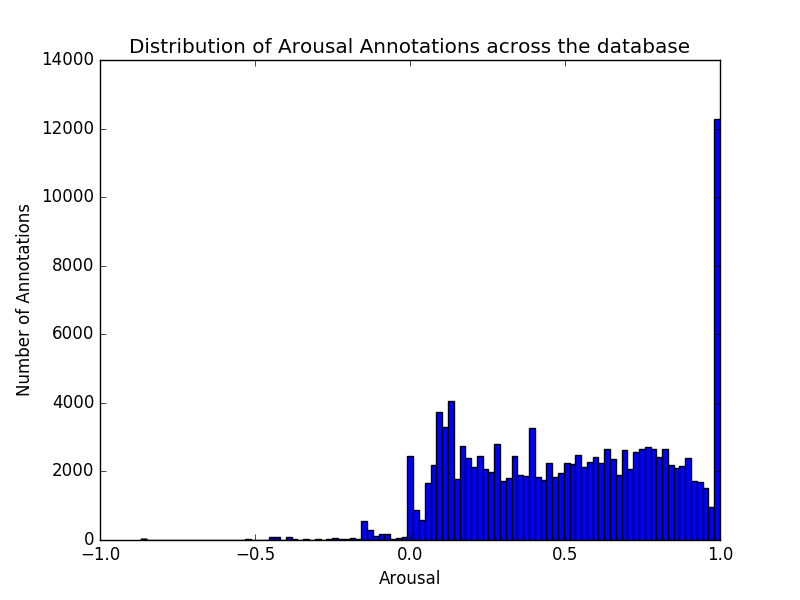}
\caption{Histogram of valence and arousal annotations of the part of Aff-Wild annotated for the seven basic expressions.}
\label{hist_va_bas_expr}
\vskip -0.3cm
\end{figure}

Figure \ref{hist_va_bas_expr} provides a histogram of the valence and arousal
values in the annotated part of Aff-Wild.

\subsection{Database partition sets}\label{sets}

\subsubsection{AU Sets}\label{sets}

The total set that we kept was the 64 videos consisting of $139,298$ number of frames, that had at least one AU activated, plus other $40,702$ that did not have any AU activated. So our total set had $180,000$ frames. We partitioned this set into training, validation and test sets. The partitioning was done in a subject independent manner, meaning that if a video in one set contained one person that was also present in other videos, then all videos containing this person should be added to the same set. 
The resulting training set consisted of $38$ videos and $107,661$ frames, the validation set consisted of $8$ videos and $23,134$ frames and the test set consisted of $18$ videos and $49,205$ frames.

\subsubsection{Basic Expression Sets}\label{sets2}

Here, we kept $55$ videos consisting of $115,640$ number of frames, that belonged to one of the basic expressions. We partitioned this set into training, validation and test sets. The partitioning was done again in a subject independent manner. 
The resulting training set consisted of $30$ videos and $67,525$ frames, the validation set consisted of $10$ videos and $20,675$ frames and the test set consisted of $15$ videos and $27,440$ frames.

\section{Pre-processing and Annotation}

\subsection{Database pre-processing}

Each of those $64$ and $55$ videos had a different frames per second (fps) rate, close or equal to 30. We converted all the videos to be in MP4 format and have 30 fps.

\subsection{Annotation tool}

We developed our own annotation software that enabled us to annotate each AU independently and frame-by-frame for each video.
For the annotation of the seven basic expressions, the same tool has been used; the only difference was the annotation tags, where the seven emotion categories were used instead of the eight AUs.

The expert-annotator could either select a time range, or some specific frames, then watch the corresponding time instances and select the AU(s)/basic expressions he/she wanted to annotate and finally perform the annotation. It should also be added that the annotation tool  also provided the ability to show the inserted annotation, while displaying a respective video. This was used for annotation verification in a post-processing step.

\subsection{Annotation procedure}

In total, three expert human coders performed the labelling, independently from each other. 
For annotating AUs, the whole video was watched first, so that the annotator could spot the most important parts, i.e. the parts in which the person reacted the most. Then, the annotation for each Action Unit was performed, by having the annotator watch the video again and select the frames where the AU was present. This process was repeated for each Action Unit.

For annotating the primary expressions, before starting the annotation of each video, the experts watched the whole video so as to know what to expect regarding the emotions being displayed in the video. Then the annotation was performed for each expression at a time. The experts took into account the fact that the seven emotion categories are not mutually exclusive (as in the case of AUs), meaning that once a frame was labelled as, e.g., happy, this frame could not be further annotated. Each frame belonged to only one specific emotion category.

\subsection{Post-processing}

Every expert-annotator watched all videos for a second
time, in order to verify that the recorded annotations
were in accordance with the shown emotions in the videos,
or to change the annotations accordingly. In this way, a further
validation of annotations was achieved.

After the annotations have been validated by the annotators,
a final annotation selection step followed.
The 64 videos contained in total $234,000$ number of frames. However, not all frames had at least one AU activated. For the AUs that were activated, the agreement between the coders was not always 100\%. That is why we decided to keep only the AU annotations in frames where all three experts agreed on the activation of an AU. In the primary expression case, experts should have a 100\% agreement in their annotations.

\subsection{Face detection}

For detecting the faces \cite{avrithis2000broadcast} in all videos, the FFLD2 Face Detector \cite{mathias2014face} was used.
The resulting facial images were resized to resolution of $96 \times 96 \times 3$ or $32 \times 32 \times 3$ , with their
intensity values being normalized to the range $[-1, 1]$.

\section{The Deep Neural Architectures}
 
In this Section, at first we describe the GANs that we adapted and used, and then present the CNN-RNN architecture, followed by the adopted error minimization criteria.

\subsection{GANs for image generation, but also joint classification and regression}\label{gans}

Here we aim at using GANs for semi-supervised learning, where the discriminator also serves as a classifier (AU detection) and regressor (VA estimation). For a semi-supervised learning problem with $k$ classes (in our case $k=2+8=10$), the discriminator has $k+1$ outputs; the last output corresponds to the fake examples, originating from the generator of the GAN (let us call it fake class). We tested two different configurations for the generator and the discriminator.

The first configuration was based on the SSGAN. The generator of this configuration is shown in Table \ref{gen_ssgan}. A 100 dimensional-vector Z is sampled from a uniform distribution in the range $[-1,1]$ and is reshaped into a 4-dimensional tensor $[1, 1, 1, 100]$, which is then passed through a series of four fractionally-strided convolutions (denoted as conv2d transpose);an image with resolution $32 \times 32 \times 3$ is generated in the end. Table \ref{disc_ssgan} shows the architecture of the corresponding discriminator. 

\begin{table}[h]
\centering
\caption{Configuration 1: Generator network of our semi-supervised GAN; output is a $32 \times 32 \times 3$ image}
\label{gen_ssgan}
\scalebox{1}{
\begin{tabular}{|c|c|c|c|}
\hline
Layer & filter  & stride & padding \\
\hline
conv2d transpose 1 & [2, 2, 100, 384]  & [1, 1, 1, 1] & 'SAME'  \\
batch normalization &&&\\
relu &&&\\
\hline
conv2d transpose 2  & [4, 4, 384, 128]  & [1, 2, 2, 1] & 'SAME'  \\
batch normalization &&&\\
relu &&&\\
\hline
conv2d transpose 3  & [4, 4, 128, 64]  & [1, 2, 2, 1] & 'SAME'  \\
batch normalization&&& \\
relu &&&\\
\hline
conv2d transpose 4  & [6, 6, 64, 3]  & [1, 2, 2, 1] & 'SAME'  \\
tanh &&&\\
\hline
\end{tabular}}
\end{table}

\begin{table}[h]
\centering
\caption{Configuration 1: Discriminator network of our semi-supervised GAN; input is a $32 \times 32 \times 3$ image}
\label{disc_ssgan}
\scalebox{0.85}{
\begin{tabular}{|c|c|c|c|c|}
\hline
Layer & filter  & stride & padding & no of units \\
\hline
conv2d 1 & [5, 5, 3, 64]  & [1, 2, 2, 1] & 'SAME' & \\
batch normalization &&&&\\
leaky relu &&&&\\
\hline
conv2d 2  & [5, 5, 64, 128]  & [1, 2, 2, 1] & 'SAME' & \\
batch normalization &&&&\\
leaky relu &&&&\\
\hline
conv2d 3  & [5, 5, 128, 256]  & [1, 2, 2, 1] & 'SAME' & \\
batch normalization&&&& \\
leaky relu &&&&\\
\hline
fully connected  &&&&2+8+1\\
sigmoid on AUs &&&& \\
sigmoid on Fake &&&& \\
\hline
\end{tabular}}
\end{table}

The second configuration was a modification of the DCGAN. The generator of this configuration is shown in Table \ref{gen_dcgan}. A $100$ dimensional-vector Z is sampled from a uniform distribution in range $[-1,1]$, projected and reshaped into a 4-dimensional tensor $[1, 6, 6, 1024]$, which is then passed through a series of four fractionally-strided convolutions and at the end an image with resolution $96 \times 96 \times 3$ is generated. Table \ref{discr_dcgan} shows the architecture of the corresponding discriminator. We also tested the same configuration for the generator and the discriminator networks, but instead of using $5 \times 5$ filters, we used $7 \times 7$.

\begin{table}[h]
\centering
\caption{Configuration 2: Generator network of our semi-supervised GAN; output is a $96 \times 96 \times 3$ image}
\label{gen_dcgan}
\scalebox{0.85}{
\begin{tabular}{|c|c|c|c|c|}
\hline
Layer & filter  & stride & padding & no of units \\
\hline
fully connected  &&&&1*6*6*1024\\
batch normalization&&&& \\
relu &&&&\\
\hline
conv2d transpose 1 & [5, 5, 1024, 512]  & [1, 2, 2, 1] & 'SAME' & \\
batch normalization &&&&\\
relu &&&&\\
\hline
conv2d transpose 2  & [5, 5, 512, 256]  & [1, 2, 2, 1] & 'SAME' & \\
batch normalization &&&&\\
relu &&&&\\
\hline
conv2d transpose 3  & [5, 5, 256, 128]  & [1, 2, 2, 1] & 'SAME' & \\
batch normalization&&&& \\
relu &&&&\\
\hline
conv2d transpose 4  & [5, 5, 128, 3]  & [1, 2, 2, 1] & 'SAME' & \\
tanh &&&&\\
\hline
\end{tabular}}
\end{table}

\begin{table}[h]
\centering
\caption{Configuration 2: Discriminator network of our semi-supervised GAN; input is a $96 \times 96 \times 3$ image}
\label{discr_dcgan}
\scalebox{0.85}{
\begin{tabular}{|c|c|c|c|c|}
\hline
Layer & filter  & stride & padding & no of units \\
\hline
conv2d 1 & [5, 5, 3, 64]  & [1, 2, 2, 1] & 'SAME' & \\
batch normalization &&&&\\
leaky relu &&&&\\
\hline
conv2d 2  & [5, 5, 64, 128]  & [1, 2, 2, 1] & 'SAME' & \\
batch normalization &&&&\\
leaky relu &&&&\\
\hline
conv2d 3  & [5, 5, 128, 256]  & [1, 2, 2, 1] & 'SAME' & \\
batch normalization&&&& \\
leaky relu &&&&\\
\hline
conv2d 4  & [5, 5, 256, 512]  & [1, 2, 2, 1] & 'SAME' & \\
batch normalization&&&& \\
leaky relu &&&&\\
\hline
fully connected  &&&&2+8+1\\
sigmoid on AUs &&&& \\
sigmoid on Fake &&&& \\
\hline
\end{tabular}}
\end{table}

Note that in both configurations there are $11$ outputs in the discriminator, while the output activation function is linear for VA, sigmoid for the eight AUs and sigmoid for the fake class. 
For comparison purposes, we tested the above two configurations with the discriminator being only classifier (of AUs), or only regressor (VA estimation), apart from predicting the fake class.

\subsection{CNN-RNN for multi-task learning: joint  classification and regression}\label{cnn-rnn_va_expr}

\begin{figure}[h]
    \centering
    \includegraphics[scale=0.27 ]{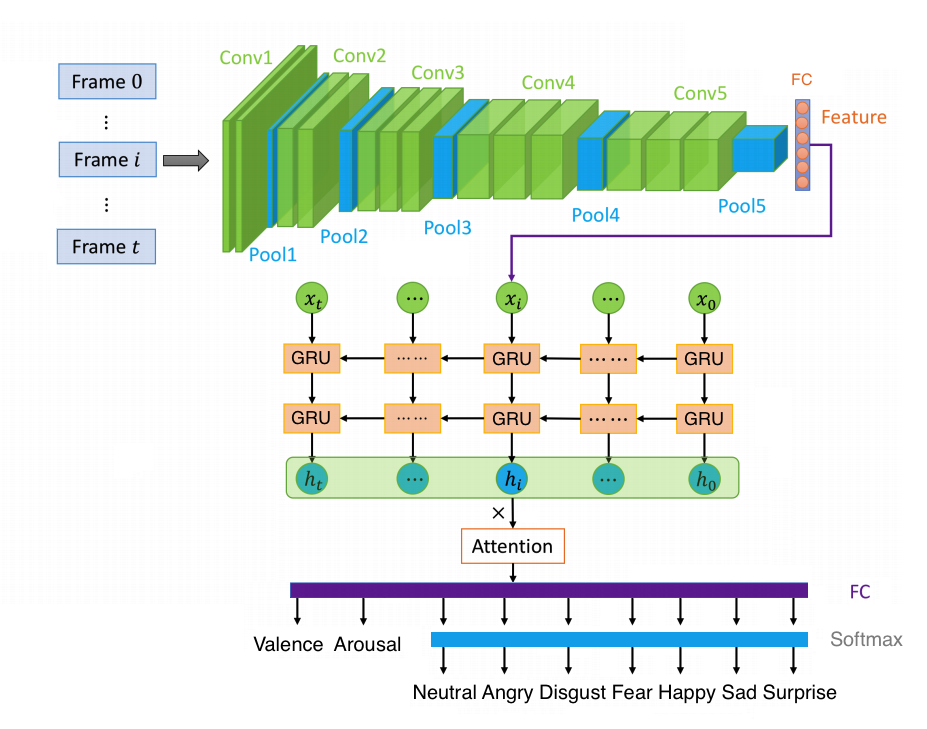}
    \caption{The multi-task learning model: a VGGFACE-GRU-attention model predicting VA and Seven Basic Expressions}
    \label{va-expr-model}
\end{figure}

Here, we aim at multi-task learning of VA and seven basic expressions. The architecture that we used was based on the AffWildNet \cite{kollias2,kollias3}, which is a VGGFACE-GRU \cite{parkhi2015deep,chung2014empirical,kollias7,kollias14} network that provided the best results in the Aff-Wild database for VA estimation. We have modified the architecture so as to include an attention layer on top of the RNN layer and 9 outputs, with the output activation function being linear for VA and softmax for the 7 basic expressions. Figure \ref{va-expr-model} shows in more detail the architecture of the model being used.

\subsection{Objective Functions \& Training implementation details}

\subsubsection{Case with Valence-Arousal \& Action Units}

The loss function used for training the GANs was: 
\vskip -0.5cm

\begin{equation} \label{eq_1}
\mathcal{L}_{total} = \mathcal{L}_{gen} + \mathcal{L}_{discr} = \mathcal{L}_{gen} + \mathcal{R}_{images} + \mathcal{F}_{images},
\end{equation}

\noindent
where $\mathcal{L}_{gen}$ is the loss of the generator, $\mathcal{R}_{images}$ and  $\mathcal{F}_{images}$ are the losses of the discriminator when presented at input with real and fake images, respectively.

For the generator loss, a reconstruction loss with an annealed weight is applied as an auxiliary loss to help the generator get rid of the initial local minimum:

\vskip -0.3cm

\begin{equation} \label{eq_2}
\mathcal{L}_{gen} = \log x + w_{r} * \mathcal{L}_{\delta}(real-fake),
\end{equation}

where $logx$ is the logarithm of the output, $x$, of the fake class of the discriminator when  presented with a fake image, $w_r$ is the annealed weight with initial value $1$ and $\mathcal{L}_{\delta}(real-fake)$ is the huber loss \cite{huber1964robust} with $\delta = 1$ between the real image and the generated fake image.

During training, for a real image the corresponding label for the fake class was 0. For a fake image, label smoothing is applied and the resulting label has a value of 0.01 for valence, arousal and the eight AUs and 0.9 for the fake class. The loss for real images was defined as:

\vskip -0.3cm

\begin{equation} \label{eq_3}
\mathcal{R}_{images} = \mathcal{R}_{VA} + \mathcal{R}_{AU} + \mathcal{R}_{fake} , 
\end{equation}

where $\mathcal{R}_{i}$ denotes the loss of class $i \in \{VA,AU,fake\} $. Note that the loss for fake images was similar.

For $\mathcal{L}_{VA}$ loss, two different losses were tested.
The first loss function was based on the Concordance Correlation Coefficient (CCC) and was defined as:

\begin{equation} \label{eq_4}
\mathcal{L}_{VA} = 1 - \frac{\rho_a + \rho_v}{2},
\end{equation}

where $\rho_a$ and $\rho_v$ is the CCC for arousal and valence, respectively and was defined as follows  :

\begin{equation} \label{eq_5}
\rho_c = \frac{2 s_{xy}}{s_x^2 + s_y^2 + (\bar{x} - \bar{y})^2},
\end{equation}

\noindent
where $s_x$ and $s_y$ are the variances of the valence/arousal labels and predicted values respectively, $\bar{x}$ and $\bar{y}$ are the corresponding mean values,  $s_{xy}$ is the respective covariance value and $c$ can be either $a$ (stands for arousal) or $v$ (stands for valence).

The second loss function was the usual Mean Squared Error (MSE). We calculated the MSE for valence and for arousal and $\mathcal{L}_{VA}$ was their average.

For $\mathcal{L}_{AU}$ loss, we used the cross entropy loss, averaged across all eight AUs. 
For the fake class, we used the cross entropy loss.

We used separate learning rates for the generator and the discriminator. The generator's learning rate was constantly set to  $10^{-4}$ and the discriminator's to $10^{-5}$. To avoid the fast convergence of the discriminator network, we updated the generator network more frequently. For configuration 1, the discriminator was updated five times before the generator, whereas for configuration 2 the discriminator was updated twice. We further applied gradient clipping with value 20, so as to stabilize training. All models were trained using the Adam optimizer with $\beta_1 = 0.5$ and $\beta_2 = 0.999$, with a mini-batch size of 64. In the LeakyReLU, the slope of the leak was set to 0.2 in all models.

\subsubsection{Case with Valence-Arousal \& Basic Expressions}\label{loss_va_au}

The loss function used for training the networks was: 

\vskip -0.3cm

\begin{equation} \label{eq:1}
\mathcal{L}_{total} = \alpha * \mathcal{L'}_{VA} + \beta * \mathcal{L}_{basic\_expressions},
\end{equation}

\noindent
where $\mathcal{L'}_{VA}$ is the loss function for valence and arousal, $\mathcal{L}_{basic\_expressions}$ is the loss function for the seven basic expressions, $\alpha$ and $\beta$ are constants with values in $[0,1]$. Note that if $\alpha=0$ and $\beta=1$, or $\alpha=1$ and $\beta=0$, then the network acts as only a classifier or only a regressor, respectively.

Different loss functions were used. $\mathcal{L'}_{VA}$ was either $\mathcal{L}_{VA}$ of Eq. \ref{eq_4} or MSE. $\mathcal{L}_{basic\_expressions}$ was either cross entropy or MSE. In the basic expressions case, when the loss function was the MSE, two approaches were tried: either passing the output of the softmax (i.e., the probabilities of the emotion categories), or the output of the fully connected layer before the softmax.

The network was trained either end-to-end, or we kept the CNN weights fixed and trained the RNN, attention and output layers.
For network training, we utilized the Adam optimizer; the batch size was set to 800 (consisting of 10 different sequences, each having 80 consecutive frames), the attention length was chosen to be 32 and the learning rate was set to 0.001 or 0.0001.
The platform used was Tensorflow.

\subsection{The evaluation criteria}\label{criteria}


The criteria that were considered for evaluating the performance of the networks were:
\begin{itemize}
    \item[i)] for AUs and basic expressions: total accuracy, weighted and macro f1 score 
    \item[ii)] for VA: CCC
\end{itemize}
Taking into account the imbalanced distribution of the AUs and basic expressions, we did not want to evaluate the models only on the accuracy metric, as it is sensitive to bias and not really effective for imbalanced data. Also the macro f1 score (unweighted average of f1 scores of all AUs) does not account for imbalanced classes. That is why we also considered the weighted f1 score (weighted average of f1 scores of all AU).

\section{Experimental Results}\label{experiments}

\subsection{Case with Valence-Arousal \& Action Units}


\subsubsection{Results on Performance of Classifier and/or Regressor}

Here, we examine the performance of the GANs described in Section \ref{gans}, using different loss functions for the discriminator network, when: predicting only the VA; classifying only into the basic expressions; serving as both regressor and classifier. Table \ref{gan_au_va} shows that a better performance is obtained when the discriminator classifies only the AUs, or  performs only VA estimation, than when it jointly predicts VA and AUs. The Table also shows that the discriminator has a better performance in predicting only the VA, when the MSE is used as loss function, than  when the loss function is based on CCC.

\begin{table}[h]
\caption{Performance of GAN's Discriminator when trained with different loss functions and when serving as a regressor-only, a classifier-only or a regressor-classifier}
\label{gan_au_va}
\centering
\scalebox{0.9}{
\begin{tabular}{ |c|c|c|c|c|c|c|  }
 \hline
 \begin{tabular}{@{}c@{}} Discriminator \\ that also \\ classifies \end{tabular} & Loss function & \begin{tabular}{@{}c@{}} CCC-V \\ CCC-A \end{tabular}  & \begin{tabular}{@{}c@{}} F1 Score: \\ weighted \\ macro \end{tabular} & \begin{tabular}{@{}c@{}} Total  Accuracy \end{tabular}  \\
\hhline{|=|=|=|=|=|}
only VA & CCC based & \begin{tabular}{@{}c@{}} 0.440 \\ 0.370 \end{tabular}  &  - & -     \\
\hline
\textbf{only VA} & \textbf{MSE}  & \begin{tabular}{@{}c@{}} \textbf{0.484} \\ \textbf{0.384} \end{tabular}  &  - & -     \\
\hline
\textbf{only AU} &  \textbf{cross entropy} & -  &  \begin{tabular}{@{}c@{}} \textbf{0.404} \\ \textbf{0.40} \end{tabular} &  \begin{tabular}{@{}c@{}} \textbf{0.824}  \end{tabular}  \\
\hline
 \begin{tabular}{@{}c@{}}VA \\ + \\ AU  \end{tabular} & \begin{tabular}{@{}c@{}} CCC based \\ + \\  cross entropy \end{tabular}  & \begin{tabular}{@{}c@{}} 0.447 \\ 0.374 \end{tabular} & \begin{tabular}{@{}c@{}} 0.328 \\ 0.330 \end{tabular}  &  \begin{tabular}{@{}c@{}} 0.661  \end{tabular}  \\
\hline
 \begin{tabular}{@{}c@{}}VA \\ + \\ AU  \end{tabular} & \begin{tabular}{@{}c@{}} MSE \\ + \\  cross entropy \end{tabular}  & \begin{tabular}{@{}c@{}} 0.451 \\ 0.333 \end{tabular} & \begin{tabular}{@{}c@{}} 0.332 \\ 0.347 \end{tabular}  &  \begin{tabular}{@{}c@{}} 0.667  \end{tabular}  \\
\hline
\end{tabular}
}
\end{table}


\subsubsection{Image Generation}

\begin{figure}[h]
\centering
\begin{tabular}{ c  }
\includegraphics[height=0.95cm]{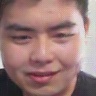}\includegraphics[height=0.95cm]{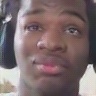}\includegraphics[height=0.95cm]{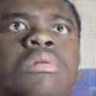}\includegraphics[height=0.95cm]{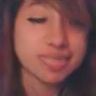}\includegraphics[height=0.95cm]{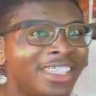}\includegraphics[height=0.95cm]{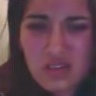}\includegraphics[height=0.95cm]{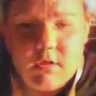}\includegraphics[height=0.95cm]{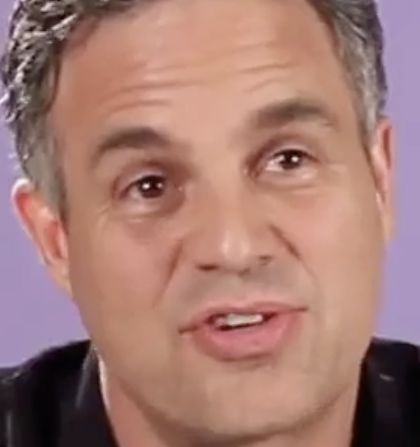}\\
\includegraphics[height=0.95cm]{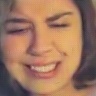}\includegraphics[height=0.95cm]{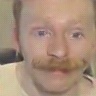}\includegraphics[height=0.95cm]{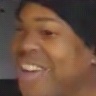}\includegraphics[height=0.95cm]{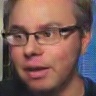}\includegraphics[height=0.95cm]{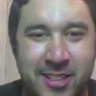}\includegraphics[height=0.95cm]{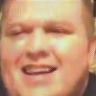}\includegraphics[height=0.95cm]{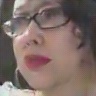}\includegraphics[height=0.95cm]{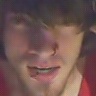}\\
 \includegraphics[height=0.95cm]{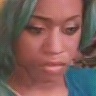}\includegraphics[height=0.95cm]{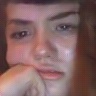}\includegraphics[height=0.95cm]{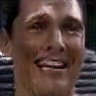}\includegraphics[height=0.95cm]{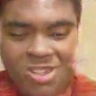}\includegraphics[height=0.95cm]{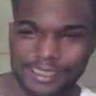}\includegraphics[height=0.95cm]{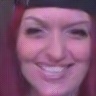}\includegraphics[height=0.95cm]{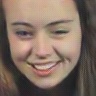}\includegraphics[height=0.95cm]{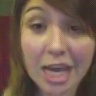}\\
\includegraphics[height=0.95cm]{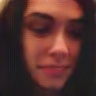}\includegraphics[height=0.95cm]{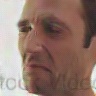}\includegraphics[height=0.95cm]{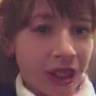}\includegraphics[height=0.95cm]{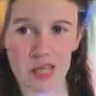}\includegraphics[height=0.95cm]{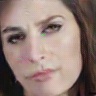}\includegraphics[height=0.95cm]{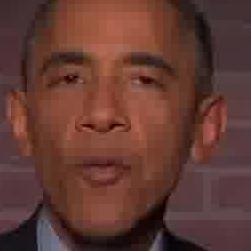} 
\end{tabular}
\caption{Images with all subjects appearing in the training set of the part of Aff-Wild that was annotated for AUs}
\label{all_identities}
\end{figure}

\begin{figure}[h]
\centering
\begin{tabular}{ c  }

\includegraphics[height=0.95cm]{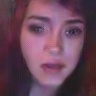}\includegraphics[height=0.95cm]{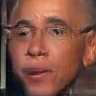}\includegraphics[height=0.95cm]{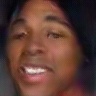}\includegraphics[height=0.95cm]{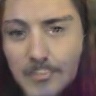}\includegraphics[height=0.95cm]{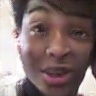}\includegraphics[height=0.95cm]{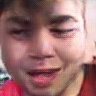}\includegraphics[height=0.95cm]{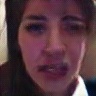}\includegraphics[height=0.95cm]{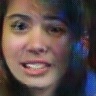}\includegraphics[height=0.95cm]{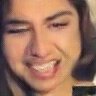}\\

\includegraphics[height=0.95cm]{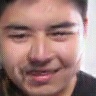}\includegraphics[height=0.95cm]{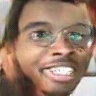}\includegraphics[height=0.95cm]{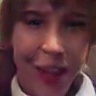}\includegraphics[height=0.95cm]{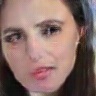}\includegraphics[height=0.95cm]{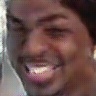}\includegraphics[height=0.95cm]{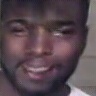}\includegraphics[height=0.95cm]{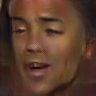}\includegraphics[height=0.95cm]{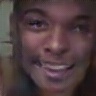}\includegraphics[height=0.95cm]{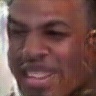}\\

\includegraphics[height=0.95cm]{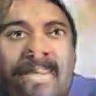}\includegraphics[height=0.95cm]{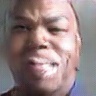}\includegraphics[height=0.95cm]{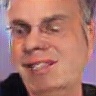}\includegraphics[height=0.95cm]{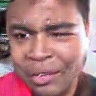}\includegraphics[height=0.95cm]{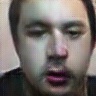}\includegraphics[height=0.95cm]{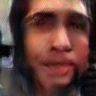}\includegraphics[height=0.95cm]{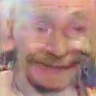}
\includegraphics[height=0.95cm]{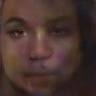}\includegraphics[height=0.95cm]{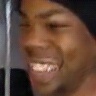}\\

\includegraphics[height=0.95cm]{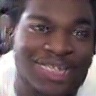}\includegraphics[height=0.95cm]{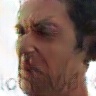}\includegraphics[height=0.95cm]{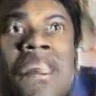}\includegraphics[height=0.95cm]{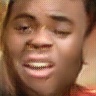}\includegraphics[height=0.95cm]{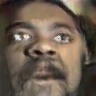}\includegraphics[height=0.95cm]{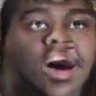}\includegraphics[height=0.95cm]{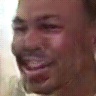}\includegraphics[height=0.95cm]{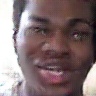}\includegraphics[height=0.95cm]{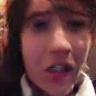}\\

\includegraphics[height=0.95cm]{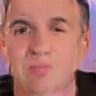}\includegraphics[height=0.95cm]{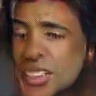}\includegraphics[height=0.95cm]{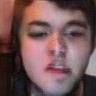}\includegraphics[height=0.95cm]{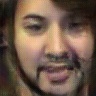}\includegraphics[height=0.95cm]{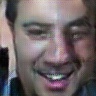}\includegraphics[height=0.95cm]{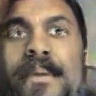}\includegraphics[height=0.95cm]{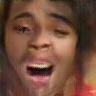}\includegraphics[height=0.95cm]{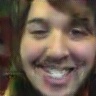}\includegraphics[height=0.95cm]{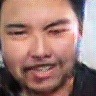}\\

\includegraphics[height=0.95cm]{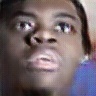}\includegraphics[height=0.95cm]{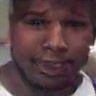}\includegraphics[height=0.95cm]{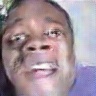}\includegraphics[height=0.95cm]{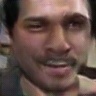}\includegraphics[height=0.95cm]{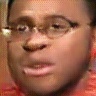}\includegraphics[height=0.95cm]{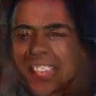}\includegraphics[height=0.95cm]{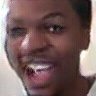}\includegraphics[height=0.95cm]{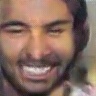}\includegraphics[height=0.95cm]{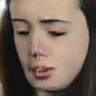}\\

\includegraphics[height=0.95cm]{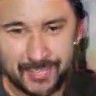}\includegraphics[height=0.95cm]{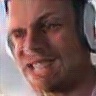}\includegraphics[height=0.95cm]{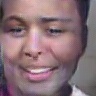}\includegraphics[height=0.95cm]{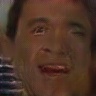}\includegraphics[height=0.95cm]{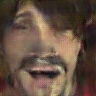}\includegraphics[height=0.95cm]{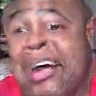}\includegraphics[height=0.95cm]{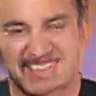}\includegraphics[height=0.95cm]{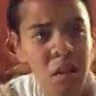}\includegraphics[height=0.95cm]{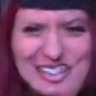}\\

\includegraphics[height=0.95cm]{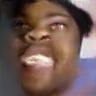}\includegraphics[height=0.95cm]{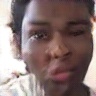}\includegraphics[height=0.95cm]{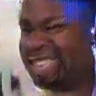}\includegraphics[height=0.95cm]{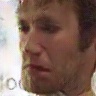}\includegraphics[height=0.95cm]{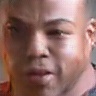}\includegraphics[height=0.95cm]{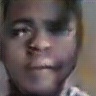}\includegraphics[height=0.95cm]{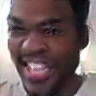}\includegraphics[height=0.95cm]{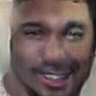}\includegraphics[height=0.95cm]{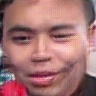}\\

\includegraphics[height=0.95cm]{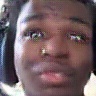}\includegraphics[height=0.95cm]{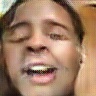}\includegraphics[height=0.95cm]{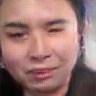}\includegraphics[height=0.95cm]{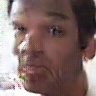}\includegraphics[height=0.95cm]{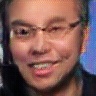}\includegraphics[height=0.95cm]{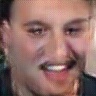}\includegraphics[height=0.95cm]{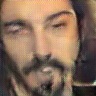}\includegraphics[height=0.95cm]{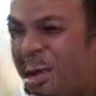}\includegraphics[height=0.95cm]{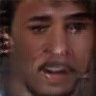}\\

\includegraphics[height=0.95cm]{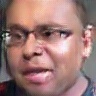}\includegraphics[height=0.95cm]{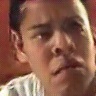}\includegraphics[height=0.95cm]{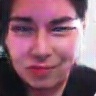}\includegraphics[height=0.95cm]{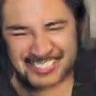}\includegraphics[height=0.95cm]{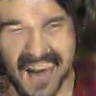}\includegraphics[height=0.95cm]{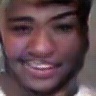}\includegraphics[height=0.95cm]{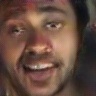}\includegraphics[height=0.95cm]{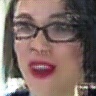}\includegraphics[height=0.95cm]{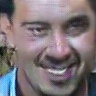}\\

\end{tabular}
\caption{Generated images from GAN}
\label{synthetic}
\end{figure}

Here we qualitatively evaluate the quality of image synthesis by the generator of the GANs. Figure \ref{all_identities} shows all subjects appearing in the training set of the Aff-Wild videos that have been annotated for AUs. Figure \ref{synthetic} shows examples of generated images.

 It can be observed that the generator has: i) created new faces and ii) created new expressions or modified some attributes (i.e., gender, hair, skin or eye color) of a subject appearing in the training set. Such example cases are a girl having moustache or a guy having girl's hair, or a subject wearing glasses (but did not wear in the real images), or a guy having now beard, or a girl having different hair color. Although the training set did not contain a lot of different subjects, the generator was able to adequately learn the in-the-wild nature of Aff-Wild (with great variability in expressed emotions, with different head poses and illumination conditions and with occlusions), 'transferring' this to the generated images.

\subsection{Case with Valence-Arousal \& Basic Expressions}

Here, we evaluate the performance, on the test set, of the deep neural architecture described in Section \ref{cnn-rnn_va_expr}. 
At first, we compare the performance of this architecture when trained with different loss functions described in Section \ref{loss_va_au} and with different learning rates. We should mention that for these experiments, we kept the values of $\alpha$ and $\beta$ fixed, at value 0.5.

Table \ref{all} shows the obtained performance in VA and basic expression estimation, using different loss functions and learning rate values. It can be seen that the best performance, under all three evaluation criteria (CCC, Total Accuracy, F1 Score) was achieved when the valence arousal loss was CCC based, the basic expressions loss was cross entropy and the learning rate was $10^{-3}$.

Next, we used the above best performing network with different values for $\alpha$ and $\beta$ in the loss function in Eq. \ref{eq:1}. It can be seen, from Table \ref{diff_a_b}, that when $\alpha = \beta = 0.5$ (as in the experiments of Table \ref{all}) the network had the best performance according to all three evaluation criteria. This performance was better than when training the network to only predict valence-arousal (which is the case when $\alpha=1$ and $\beta=0$), or to only classify in the seven basic expressions (which is the case when $\alpha=0$ and $\beta=1$). This shows that the multi-task learning provided an improved performance compared to both single learning cases, by taking advantage of the relations existing in the learned tasks.

\begin{table}[h]
\caption{Obtained performances when comparing best model from Table \ref{all} with different values for $\alpha$ and $\beta$ in the loss function in Eq. \ref{eq:1}}
\label{diff_a_b}
\centering
\scalebox{0.9}{
\begin{tabular}{ |c|c|c|c|  }
 \hline
 \begin{tabular}{@{}c@{}} Network \\ with ($\alpha$ , $\beta$) \end{tabular} & \begin{tabular}{@{}c@{}} CCC: \\ V-A \end{tabular}  & \begin{tabular}{@{}c@{}} Total \\ Accuracy \end{tabular} & \begin{tabular}{@{}c@{}} F1: \\ Weighted - Unweighted \end{tabular}  \\
\hhline{|=|=|=|=|}
\begin{tabular}{@{}c@{}} (0 , 1) $\equiv$  only expr.  \end{tabular} & - & 0.494  &  \begin{tabular}{@{}c@{}} 0.488 - 0.415 \end{tabular}  \\
\hline
\begin{tabular}{@{}c@{}} (1 , 0) $\equiv$  only VA\end{tabular} & \begin{tabular}{@{}c@{}} 0.579 - 0.409 \end{tabular} & -  & -  \\
\hline
\begin{tabular}{@{}c@{}} (0.25 , 0.75)  \end{tabular} & \begin{tabular}{@{}c@{}} 0.556 - 0.419 \end{tabular} & 0.547  &  \begin{tabular}{@{}c@{}} 0.542 - 0.452 \end{tabular}  \\
\hline
\begin{tabular}{@{}c@{}} (0.75 , 0.25)  \end{tabular} & \begin{tabular}{@{}c@{}} 0.589 - 0.424 \end{tabular} & 0.527  &  \begin{tabular}{@{}c@{}} 0.514 - 0.422 \end{tabular}  \\
\hline
\begin{tabular}{@{}c@{}} \textbf{(0.5 , 0.5)} \end{tabular} & \begin{tabular}{@{}c@{}} \textbf{0.616} - \textbf{0.510} \end{tabular} & \textbf{0.645}  &  \begin{tabular}{@{}c@{}} \textbf{0.643} - \textbf{0.514} \end{tabular}  \\
\hline
\begin{tabular}{@{}c@{}} (0.75 , 0.75)  \end{tabular} & \begin{tabular}{@{}c@{}} 0.585 - 0.453 \end{tabular} & 0.561  &  \begin{tabular}{@{}c@{}} 0.555 - 0.476 \end{tabular}  \\
\hline
\begin{tabular}{@{}c@{}} (1 , 1) \end{tabular} & \begin{tabular}{@{}c@{}} 0.521 - 0.356 \end{tabular} & 0.513  & \begin{tabular}{@{}c@{}} 0.501 - 0.411 \end{tabular}   \\
\hline
\end{tabular}
}
\end{table}

\begin{table*}[h]
\caption{Obtained values when testing different loss functions and learning rates}
\label{all}
\centering
\begin{tabular}{ |c|c|c|c|c|c|c|c| }
 \hline
 \multicolumn{3}{|c|}{Parameters} & \multicolumn{2}{c|}{CCC} & \multicolumn{1}{c|}{Accuracy} & \multicolumn{2}{c|}{F1 Score}  \\
 \hline
 VA-Loss & Basic Expressions-Loss & Learning Rate  & Valence & Arousal & & Unweighted & Weighted    \\
\hhline{|=|=|=|=|=|=|=|=|}
CCC based & Cross Entropy & $10^{-4}$ &0.510 &0.445 & 0.644  & 0.619  & 0.481   \\
 \hline
\textbf{CCC based} & \textbf{Cross Entropy} & $\textbf{10}^{\textbf{-3}}$ &\textbf{0.616} &\textbf{0.510} & \textbf{0.645}  & \textbf{0.643}  & \textbf{0.514}  \\
\hline
CCC based & MSE before softmax & $10^{-4}$ &0.491 &0.433 & 0.588  & 0.581  & 0.497  \\
 \hline
CCC based & MSE before softmax & $10^{-3}$ &0.507 &0.441 & 0.517 & 0.501  & 0.469  \\
\hline
CCC based & MSE after softmax & $10^{-4}$ &0.462 &0.406 & 0.575 & 0.562  & 0.488  \\
 \hline
CCC based & MSE after softmax & $10^{-3}$ &0.486 &0.422 & 0.569  & 0.511  & 0.447 \\
\hline
MSE & Cross Entropy & $10^{-4}$ &0.502 &0.350 & 0.641  & 0.620  & 0.492   \\
\hline
MSE & Cross Entropy & $10^{-3}$ &0.532 &0.362 & 0.601  & 0.606  & 0.480   \\
\hline
MSE & MSE before softmax & $10^{-4}$ &0.537 &0.482 & 0.600  & 0.598  & 0.476   \\
\hline
MSE & MSE before softmax & $10^{-3}$ &0.561 &0.487 & 0.556  & 0.541  & 0.448   \\
\hline
MSE & MSE after softmax & $10^{-4}$ &0.495 &0.465 & 0.602  & 0.577  & 0.471   \\
\hline
MSE & MSE after softmax & $10^{-3}$ &0.515 &0.468 & 0.581  & 0.527  & 0.423   \\
\hline
\end{tabular}
\end{table*}

\section{CONCLUSIONS AND FUTURE WORK}

A multi-task learning for emotion recognition and for facial image generation was presented in this paper. This was made possible by annotating large parts of the Aff-Wild database in terms of Facial Action Units and in terms of the seven Basic Expressions and underlying emotions. Deep neural architectures and GANs were appropriately designed and trained using these annotated datasets. High performances have been obtained in all multi-task experiments. Moreover, enrichment of the above datasets was achieved through the image generation approach implemented by GANs. Future work will include extension of the developed multi-tasking models and datasets, so as to further improve the capabilities of the proposed systems.        

\bibliographystyle{ieee}
\bibliography{egbib}

\begin{thebibliography}{10}\itemsep=-1pt

\bibitem{avrithis2000broadcast}
Y.~Avrithis, N.~Tsapatsoulis, and S.~Kollias.
\newblock Broadcast news parsing using visual cues: A robust face detection
  approach.
\newblock In {\em 2000 IEEE International Conference on Multimedia and Expo.
  ICME2000. Proceedings. Latest Advances in the Fast Changing World of
  Multimedia (Cat. No. 00TH8532)}, volume~3, pages 1469--1472. IEEE, 2000.

\bibitem{berthelot2017began}
D.~Berthelot, T.~Schumm, and L.~Metz.
\newblock Began: boundary equilibrium generative adversarial networks.
\newblock {\em arXiv preprint arXiv:1703.10717}, 2017.

\bibitem{caruana1997multitask}
R.~Caruana.
\newblock Multitask learning.
\newblock {\em Machine learning}, 28(1):41--75, 1997.

\bibitem{chung2014empirical}
J.~Chung, C.~Gulcehre, K.~Cho, and Y.~Bengio.
\newblock Empirical evaluation of gated recurrent neural networks on sequence
  modeling.
\newblock {\em arXiv preprint arXiv:1412.3555}, 2014.

\bibitem{corneanu2016survey}
C.~Corneanu, M.~Oliu, J.~Cohn, and S.~Escalera.
\newblock Survey on rgb, 3d, thermal, and multimodal approaches for facial
  expression recognition: History, trends, and affect-related applications.
\newblock {\em IEEE transactions on pattern analysis and machine intelligence},
  2016.

\bibitem{cowie2003describing}
R.~Cowie and R.~R. Cornelius.
\newblock Describing the emotional states that are expressed in speech.
\newblock {\em Speech communication}, 40(1):5--32, 2003.

\bibitem{dalgleish2000handbook}
T.~Dalgleish and M.~Power.
\newblock {\em Handbook of cognition and emotion}.
\newblock John Wiley \& Sons, 2000.

\bibitem{ekman1977facial}
P.~Ekman and W.~V. Friesen.
\newblock Facial action coding system.
\newblock 1977.

\bibitem{ganin2014unsupervised}
Y.~Ganin and V.~Lempitsky.
\newblock Unsupervised domain adaptation by backpropagation.
\newblock {\em arXiv preprint arXiv:1409.7495}, 2014.

\bibitem{goodfellow2014generative}
I.~Goodfellow, J.~Pouget-Abadie, M.~Mirza, B.~Xu, D.~Warde-Farley, S.~Ozair,
  A.~Courville, and Y.~Bengio.
\newblock Generative adversarial nets.
\newblock In {\em Advances in neural information processing systems}, pages
  2672--2680, 2014.

\bibitem{gross2010multi}
R.~Gross, I.~Matthews, J.~Cohn, T.~Kanade, and S.~Baker.
\newblock Multi-pie.
\newblock {\em Image and Vision Computing}, 28(5):807--813, 2010.

\bibitem{hinton2015distilling}
G.~Hinton, O.~Vinyals, and J.~Dean.
\newblock Distilling the knowledge in a neural network.
\newblock {\em arXiv preprint arXiv:1503.02531}, 2015.

\bibitem{huber1964robust}
P.~J. Huber et~al.
\newblock Robust estimation of a location parameter.
\newblock {\em The annals of mathematical statistics}, 35(1):73--101, 1964.

\bibitem{kollias8}
D.~Kollias, S.~Cheng, M.~Pantic, and S.~Zafeiriou.
\newblock Photorealistic facial synthesis in the dimensional affect space.
\newblock In {\em Proceedings of the European Conference on Computer Vision
  (ECCV)}, pages 0--0, 2018.

\bibitem{kollias9}
D.~Kollias, S.~Cheng, E.~Ververas, I.~Kotsia, and S.~Zafeiriou.
\newblock Generating faces for affect analysis.
\newblock {\em arXiv preprint arXiv:1811.05027}, 2018.

\bibitem{kollias10}
D.~Kollias, G.~Marandianos, A.~Raouzaiou, and A.-G. Stafylopatis.
\newblock Interweaving deep learning and semantic techniques for emotion
  analysis in human-machine interaction.
\newblock In {\em 2015 10th International Workshop on Semantic and Social Media
  Adaptation and Personalization (SMAP)}, pages 1--6. IEEE, 2015.

\bibitem{kollias2}
D.~Kollias, M.~A. Nicolaou, I.~Kotsia, G.~Zhao, and S.~Zafeiriou.
\newblock Recognition of affect in the wild using deep neural networks.
\newblock In {\em Proceedings of the IEEE Conference on Computer Vision and
  Pattern Recognition Workshops}, pages 26--33, 2017.

\bibitem{kollias11}
D.~Kollias, A.~Tagaris, and A.~Stafylopatis.
\newblock On line emotion detection using retrainable deep neural networks.
\newblock In {\em 2016 IEEE Symposium Series on Computational Intelligence
  (SSCI)}, pages 1--8. IEEE, 2016.

\bibitem{kollias13}
D.~Kollias, A.~Tagaris, A.~Stafylopatis, S.~Kollias, and G.~Tagaris.
\newblock Deep neural architectures for prediction in healthcare.
\newblock {\em Complex \& Intelligent Systems}, 4(2):119--131, 2018.

\bibitem{kollias3}
D.~Kollias, P.~Tzirakis, M.~A. Nicolaou, A.~Papaioannou, G.~Zhao, B.~Schuller,
  I.~Kotsia, and S.~Zafeiriou.
\newblock Deep affect prediction in-the-wild: Aff-wild database and challenge,
  deep architectures, and beyond.
\newblock {\em International Journal of Computer Vision}, 127(6-7):907--929,
  2019.

\bibitem{kollias12}
D.~Kollias, M.~Yu, A.~Tagaris, G.~Leontidis, A.~Stafylopatis, and S.~Kollias.
\newblock Adaptation and contextualization of deep neural network models.
\newblock In {\em 2017 IEEE Symposium Series on Computational Intelligence
  (SSCI)}, pages 1--8. IEEE, 2017.

\bibitem{kollias4}
D.~Kollias and S.~Zafeiriou.
\newblock Aff-wild2: Extending the aff-wild database for affect recognition.
\newblock {\em arXiv preprint arXiv:1811.07770}, 2018.

\bibitem{kollias7}
D.~Kollias and S.~Zafeiriou.
\newblock A multi-component cnn-rnn approach for dimensional emotion
  recognition in-the-wild.
\newblock {\em arXiv preprint arXiv:1805.01452}, 2018.

\bibitem{kollias6}
D.~Kollias and S.~Zafeiriou.
\newblock Training deep neural networks with different datasets in-the-wild:
  The emotion recognition paradigm.
\newblock In {\em 2018 International Joint Conference on Neural Networks
  (IJCNN)}, pages 1--8. IEEE, 2018.

\bibitem{kollias14}
D.~Kollias and S.~Zafeiriou.
\newblock Exploiting multi-cnn features in cnn-rnn based dimensional emotion
  recognition on the omg in-the-wild dataset.
\newblock {\em arXiv preprint arXiv:1910.01417}, 2019.

\bibitem{kollias15}
D.~Kollias and S.~Zafeiriou.
\newblock Expression, affect, action unit recognition: Aff-wild2, multi-task
  learning and arcface.
\newblock {\em arXiv preprint arXiv:1910.04855}, 2019.

\bibitem{lucey2010extended}
P.~Lucey, J.~F. Cohn, T.~Kanade, J.~Saragih, Z.~Ambadar, and I.~Matthews.
\newblock The extended cohn-kanade dataset (ck+): A complete dataset for action
  unit and emotion-specified expression.
\newblock In {\em Computer Vision and Pattern Recognition Workshops (CVPRW),
  2010 IEEE Computer Society Conference on}, pages 94--101. IEEE, 2010.

\bibitem{mathias2014face}
M.~Mathias, R.~Benenson, M.~Pedersoli, and L.~Van~Gool.
\newblock Face detection without bells and whistles.
\newblock In {\em European Conference on Computer Vision}, pages 720--735.
  Springer, 2014.

\bibitem{pantic2005web}
M.~Pantic, M.~Valstar, R.~Rademaker, and L.~Maat.
\newblock Web-based database for facial expression analysis.
\newblock In {\em Multimedia and Expo, 2005. ICME 2005. IEEE International
  Conference on}, pages 5--pp. IEEE, 2005.

\bibitem{parkhi2015deep}
O.~M. Parkhi, A.~Vedaldi, and A.~Zisserman.
\newblock Deep face recognition.
\newblock In {\em BMVC}, volume~1, page~6, 2015.

\bibitem{plutchik1980emotion}
R.~Plutchik.
\newblock {\em Emotion: A psychoevolutionary synthesis}.
\newblock Harpercollins College Division, 1980.

\bibitem{radford2015unsupervised}
A.~Radford, L.~Metz, and S.~Chintala.
\newblock Unsupervised representation learning with deep convolutional
  generative adversarial networks.
\newblock {\em arXiv preprint arXiv:1511.06434}, 2015.

\bibitem{russell1978evidence}
J.~A. Russell.
\newblock Evidence of convergent validity on the dimensions of affect.
\newblock {\em Journal of personality and social psychology}, 36(10):1152,
  1978.

\bibitem{salimans2016improved}
T.~Salimans, I.~Goodfellow, W.~Zaremba, V.~Cheung, A.~Radford, and X.~Chen.
\newblock Improved techniques for training gans.
\newblock In {\em Advances in Neural Information Processing Systems}, pages
  2234--2242, 2016.

\bibitem{tagaris1}
A.~Tagaris, D.~Kollias, and A.~Stafylopatis.
\newblock Assessment of parkinson’s disease based on deep neural networks.
\newblock In {\em International Conference on Engineering Applications of
  Neural Networks}, pages 391--403. Springer, 2017.

\bibitem{tagaris2}
A.~Tagaris, D.~Kollias, A.~Stafylopatis, G.~Tagaris, and S.~Kollias.
\newblock Machine learning for neurodegenerative disorder diagnosis—survey of
  practices and launch of benchmark dataset.
\newblock {\em International Journal on Artificial Intelligence Tools},
  27(03):1850011, 2018.

\bibitem{tian2001recognizing}
Y.-l. Tian, T.~Kanade, and J.~F. Cohn.
\newblock Recognizing action units for facial expression analysis.
\newblock {\em Pattern Analysis and Machine Intelligence, IEEE Transactions
  on}, 23(2):97--115, 2001.

\bibitem{tu2007learning}
Z.~Tu.
\newblock Learning generative models via discriminative approaches.
\newblock In {\em Computer Vision and Pattern Recognition, 2007. CVPR'07. IEEE
  Conference on}, pages 1--8. IEEE, 2007.

\bibitem{valstar2010induced}
M.~Valstar and M.~Pantic.
\newblock Induced disgust, happiness and surprise: an addition to the mmi
  facial expression database.
\newblock In {\em Proc. 3rd Intern. Workshop on EMOTION (satellite of LREC):
  Corpora for Research on Emotion and Affect}, page~65, 2010.

\bibitem{whissel1989dictionary}
C.~Whissel.
\newblock The dictionary of affect in language, emotion: Theory, research and
  experience: vol. 4, the measurement of emotions, r.
\newblock {\em Plutchik and H. Kellerman, Eds., New York: Academic}, 1989.

\bibitem{yin2008high}
L.~Yin, X.~Chen, Y.~Sun, T.~Worm, and M.~Reale.
\newblock A high-resolution 3d dynamic facial expression database.
\newblock In {\em Automatic Face \& Gesture Recognition, 2008. FG'08. 8th IEEE
  International Conference On}, pages 1--6. IEEE, 2008.

\bibitem{yin20063d}
L.~Yin, X.~Wei, Y.~Sun, J.~Wang, and M.~J. Rosato.
\newblock A 3d facial expression database for facial behavior research.
\newblock In {\em Automatic face and gesture recognition, 2006. FGR 2006. 7th
  international conference on}, pages 211--216. IEEE, 2006.

\bibitem{kollias1}
S.~Zafeiriou, D.~Kollias, M.~A. Nicolaou, A.~Papaioannou, G.~Zhao, and
  I.~Kotsia.
\newblock Aff-wild: Valence and arousal'in-the-wild'challenge.
\newblock In {\em Proceedings of the IEEE Conference on Computer Vision and
  Pattern Recognition Workshops}, pages 34--41, 2017.

\end{thebibliography}

\end{document}